\pdfoutput=1

\documentclass[11pt]{article}

\usepackage{EMNLP2023}

\usepackage{times}
\usepackage{latexsym}
\usepackage{booktabs}
\usepackage{graphicx}
\usepackage{amsmath}
\usepackage{amssymb}
\usepackage{hyperref}
\usepackage{multirow} 
\usepackage{siunitx}  
\usepackage{todonotes}  
\usepackage{booktabs, siunitx, adjustbox} 
\usepackage[T1]{fontenc}

\usepackage[utf8]{inputenc}

\usepackage{microtype}

\usepackage{inconsolata}

\title{Context-aware Biases for Length Extrapolation}

\author{
  Ali Veisi\thanks{\enspace All authors contributed equally to this work.} \quad
  Hamidreza Amirzadeh\footnotemark[1] \quad
  Amir Mansourian\footnotemark[1] \\
  Algonet \\
  \texttt{\{ali.veisi, h.amirzadeh, amir.mansurian\}@algonetlabs.com}
}

\begin{document}
\maketitle
\begin{abstract}
Transformers often struggle to generalize to longer sequences than those seen during training—a limitation known as length extrapolation. Most existing Relative Positional Encoding (RPE) methods attempt to address this by introducing either fixed linear biases or globally learned biases, which lack the capacity to adapt to different input contexts. In this work, we propose an additive RPE, \textbf{C}ontext-\textbf{A}ware \textbf{B}iases for \textbf{L}ength \textbf{E}xtrapolation (CABLE), a method that learns token-specific, context-aware biases for each attention head in transformers. By dynamically adjusting positional biases based on the input sequence, CABLE overcomes the rigidity of fixed RPEs. When evaluated on sequences longer than originally trained with, GPT-2 Medium (334M parameters) with CABLE achieves lower perplexity than counterparts using other widely adopted positional encoding methods. Additionally, by applying CABLE to the BERT base model we improved performance in long-context retrieval tasks. Our method significantly enhances the extrapolation performance of existing RPE methods tested on the FineWeb-Edu-10B and WikiText-103 datasets. 
Our code is available at: \href{https://github.com/AlgonetLabs/Cable}{https://github.com/AlgonetLabs/Cable}.
\end{abstract}

\section{Introduction}
\label{intro}


\begin{figure}[t]
    \centering
    \includegraphics[width=\linewidth]{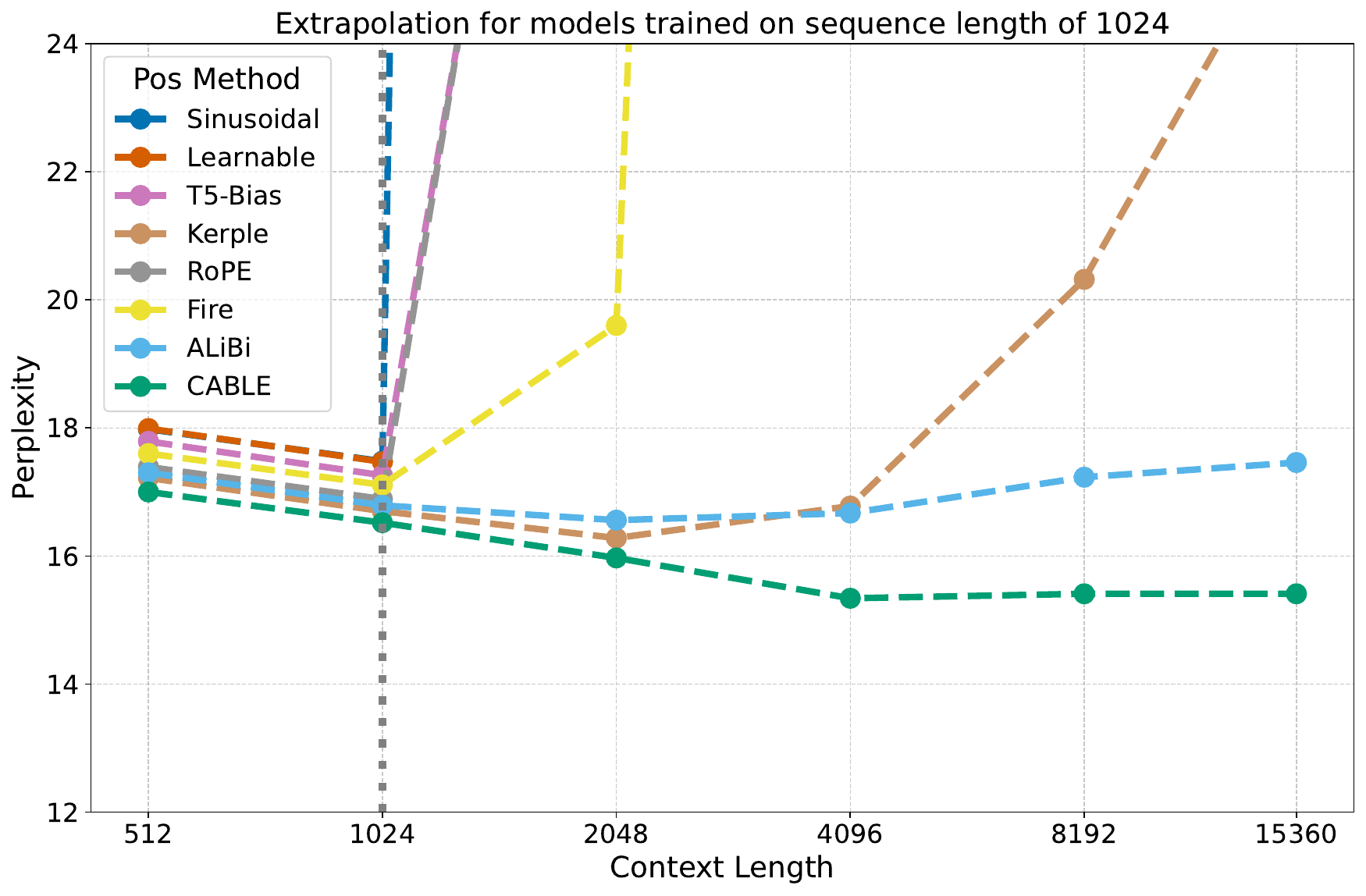}
    \caption{Next-token prediction perplexity on \texttt{FineWeb-Edu-10B} eval set with varying inference sequence lengths. The models are GPT-2 Medium trained on a sequence length of 1024 on \texttt{FineWeb-Edu-10B} train set.}
    \label{fig:pull_figure}
\end{figure}

Transformer based language models \cite{vaswani2017attention} have achieved state-of-the-art performance in many Natural Lnaguge Processing (NLP) tasks \cite{devlin-etal-2019-bert, liu2019roberta, chowdhery2023palm, team2023gemini, touvron2023llama, achiam2023gpt}. This is related to its attention mechanism that captures contextual information by considering inter-token interactions. However, in contrast to Convolutional Neural Networks (CNNs) \cite{gehring2017convolutional} and Recurrent Neural Networks (RNNs) \cite{sherstinsky2020fundamentals}, which implicitly consider positional information, transformers are shown to be position-agnostic and need position information \cite{yun2019transformers}. However, even by incorporating positional information, transformer models often experience a sharp decline in accuracy when processing inputs longer than those seen during training \cite{press2021train, anil2022exploring}. This limitation arises because training is typically performed on short sequences to mitigate the quadratic cost of attention. As a result, there is increasing interest in the length extrapolation problem—namely, a model’s ability to generalize to and accurately predict sequences longer than those encountered during training \cite{press2021train}.

Many commonly used positional encoding methods, such as Absolute Positional Encoding (APE) \cite{vaswani2017attention}, fail to generalize effectively to sequence lengths beyond those seen during training \cite{kazemnejad2024impact}. To address the length extrapolation challenge in transformers, various strategies have been proposed, including context window extension \cite{beltagy2020longformer, chen2023longlora, peng2023yarn, zhu2023pose}, memory mechanisms \cite{dai2019transformer, bulatov2022recurrent, wu2022memorizing, tworkowski2024focused}, context compression \cite{mu2024learning, tan2024lloco}, data formatting techniques \cite{shen2023positional, zhou2023algorithms}, and Relative Positional Encodings (RPE) \cite{press2021train, raffel2020exploring, su2024roformer}. Among these, RPEs have emerged as one of the most prominent and widely adopted solutions for improving length extrapolation in transformer models.

Recently, a number of RPE variants have been introduced. Rotary Positional Encoding (RoPE) \cite{su2024roformer} encodes token positions by rotating query and key vectors, while ALiBi \cite{press2021train} introduces a linear bias to attention scores. Many subsequent works have built upon these foundations, either enhancing RoPE \cite{xu2024base, peng2023yarn, chen2023extending} or refining ALiBi-style additive biases \cite{chi2022dissecting, chi2022kerple, li2023functional, gao2024mep, zhu2025rethinking}. 

In this work, based on ALiBi, we propose an additive RPE method which dynamically learns biases for tokens on each head of attention mechanism in transformers. In contrast to ALiBi that uses constant linear biases, our method, Context-aware biases for length extrapolation (CABLE), learns slopes for each head, enabling the model to create dynamic biases for each token. CABLE adds negligible time and memory burden to the conventional transformer \cite{vaswani2017attention}, while achieving better performance. As shown in Figure \ref{fig:pull_figure}, while the performance of existing positional encodings degrades with increasing sequence length, CABLE achieves even lower perplexity as sequence length increases. 
Our method is simple and easy to implement, and can be integrated into any existing transformer model easily.

Contributions of this paper are as follows:

\begin{itemize}
  \item We propose CABLE, an additive relative positional encoding method that, in contrast to existing methods, uses context-aware positional information by learning token-specific biases in each attention head. CABLE is also simple, easy to implement, and have relatively fast inference time compared to the previous methods.
  
  
  \item We evaluate our proposed method on several benchmark datasets, using GPT-2 variants for next-token prediction and BERT models for long-context retrieval. Our approach consistently outperforms existing positional encoding methods and demonstrates superior generalization to sequences longer than those seen during training.
\end{itemize}

\section{Related Work}
\label{related}


In this section, we review key approaches to positional encoding, including absolute and relative methods, as well as recent work exploring transformer models without any explicit positional encoding.

\textbf{No Positional Encoding (NoPE).} Surprisingly, \citet{haviv2022transformer} showed that decoder-only Transformers with causal attention can implicitly learn positional information without explicit encodings. \citet{kazemnejad2024impact} further supported this NoPE approach, especially in out-of-distribution (OOD) settings, suggesting that the causal mechanism alone can suffice \cite{wang2024length}. However, \citet{li2023functional} found that NoPE generally underperforms compared to models with explicit positional encodings. In a concurrent effort with our work, FoX \cite{lin2025forgetting} suggested a similar idea by not using positional encoding and instead proposed a forgetting gate. While NoPE is compatible with arbitrary sequence lengths, its performance often degrades when extrapolating far beyond training lengths.

\textbf{Absolute Positional Encoding (APE).} APE was one of the earliest approaches introduced to incorporate positional information into Transformers. \citet{vaswani2017attention} proposed both fixed (sinusoidal) and learned encodings, while \citet{gehring2017convolutional} applied learnable absolute embeddings in convolutional architectures. Later, \citet{devlin-etal-2019-bert} adopted learned absolute embeddings in BERT, adding them to token embeddings. \citet{chen2021simple} further refined APE with a decoupled attention mechanism to better separate content and positional signals. In general, APE assigns a fixed or learned vector $e_i \in \mathbb{R}^d$ to each position $i$, forming a matrix $E = [e_1, e_2, ..., e_t]^T$ that is added element-wise to token embeddings \cite{vaswani2017attention, devlin-etal-2019-bert, kiyono2021shape, likhomanenko2021cape}. A key limitation of APE methods is their poor generalization to sequence lengths beyond those seen during training, making them unsuitable for length extrapolation.

\textbf{Relative Positional Encoding (RPE).} RPE is an increasingly popular way to encode positional information for Transformers. \cite{shaw2018self} was the first to propose learning relative positional information within a clipping distance. Among the most popular methods in RPEs, is rotary positional embedding (RoPE) \cite{su2024roformer}. RoPE rotates a query and key pair vectors with an angle proportional to their relative positions before the dot product attention, which results in attention being a function of the relative distance between the tokens, capturing the relative positional information. 
One of the primary arguments for the effectiveness of RoPE—and a key reason it is widely adopted in modern LLMs—was put forth by \citet{su2024roformer}, who claimed that RoPE enables attention scores to decay as the relative distance between tokens increases. However, \citet{barbero2024round} later provided a mathematical analysis showing that this claim is flawed: attention weights under RoPE do not necessarily decay proportionally with relative query-key distances. This insight offers a possible explanation for RoPE’s limitations in length extrapolation.
In RoPE-based methods, Yarn \cite{peng2023yarn} modifies RoPE by integrating attention scaling and Neural Tangent Kernel (NTK) interpolation \cite{jacot2018neural}, and \cite{chen2023extending} extends the context window size of RoPE by interpolating positions in the range seen during training.
However, recent studies have shown that RoPE-based language models perform poorly on sequences longer than those seen during training \cite{press2021train, kazemnejad2024impact}. To address this limitation, several positional encoding methods with better length extrapolation capabilities have been proposed \cite{chen2023extending}. Among these, additive approaches have gained popularity—where a bias matrix is directly added to the pre-softmax attention logits. This design is typically intended to enforce a decay in attention weights proportional to the relative distance between query-key pairs, as shown in the following formula:
\begin{equation}
    A_{RPE}(X)=XW_Q(XW_K)^T+B
    \label{eq:additiveRPE}
\end{equation}
The bias matrix for an input sequence with $t$ tokens is $B \in \mathbb{R}^{t \times t}$, generated by a positional encoding function $b: \mathbb{N}^2 \to \mathbb{R}$, where the $(i,j)$-th entry of $B$ is given by $b(i,j)$. Naturally, different formulations of the function $b$ lead to different variants of Relative Positional Encodings (RPEs). Below are a few examples of additive RPEs that are capable of extrapolating:

    \textbf{ALiBi} \cite{press2021train}. The kernel function is defined as $b(i,j)=-r|i-j|$, where $r>0$ is a hyperparameter. ALiBi incorporates bias based on the pairwise distances into the pre-softmax attention scores. However, the function rapidly approaches the zero point \cite{chi2022kerple}, hence may not be a realistic assumption.

     \textbf{T5-bias} \cite{raffel2020exploring}. The kernel function is defined as $b(i,j) = r_{\min\{i-j,K\}}$, where $K$ is a hyperparameter and $\{{r_i}\}_{i=0}^{K}$ are learnable scalars. For positions beyond the training sequence length, the model reuses the maximum learned relative bias. While this approach allows some extrapolation, it suffers from latency issues on modern accelerators due to inefficient vectorized operations with long sequences.

     \textbf{Kerple} \cite{chi2022kerple}. The kernel function is defined as $b(i,j) = -r_1 \log(1 + r_2 |i - j|)$ in its logarithmic form and $- r_{1} |i - j|^{r_{2}}$ in its power form, where $r_1, r_2 > 0$ are learnable scalars. This approach employs a shift-invariant kernel for the bias terms.


     \textbf{Fire} \cite{li2023functional}. The kernel function is defined as $b(i,j) = f_\theta\left(\frac{\psi(i-j)}{\psi(\max\{L,i\})}\right)$, where $f_\theta$ is an MLP with $\theta$ parameters, $\psi \colon x \mapsto \log(cx + 1)$ is a monotonically increasing function and $L>0$ is a learnable scaler. This formulation allows Fire to assign more attention to distant query-key pairs—contrary to methods like ALiBi, and Kerple, which tend to focus on nearby tokens. 


\begin{figure*}[!ht]
    \centering
    \includegraphics[scale=0.46]{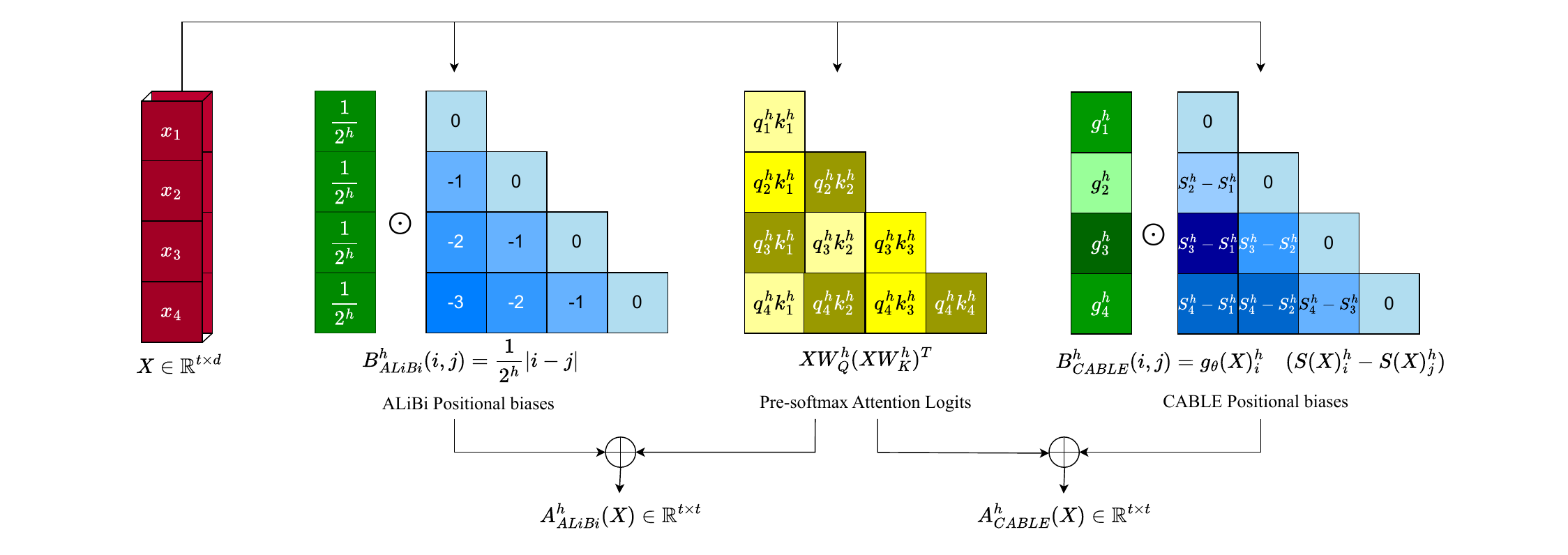}
    \caption{Comparison of how ALiBi and CABLE compute final attention scores per head. Left: ALiBi adds constant linear biases with head-specific slopes, fixed across tokens. Right: CABLE adds learned, token-specific context-aware biases and weights to the scores.}
    \label{fig:diagram}
\end{figure*}


\begin{table*}
    \centering
    \footnotesize
    \setlength{\tabcolsep}{3pt}
    \begin{tabular}{l S[table-format=2.2] S[table-format=2.2] *{9}{c}}
        \toprule
        \midrule
        \multicolumn{10}{c}{\textbf{FineWeb-Edu-10B}} \\
        \midrule 
        Sequence Length & {CABLE} & {{CABLE\textsubscript{NW}}} & {ALiBi} & {Fire} & {T5-bias} & {Kerple} & {RoPE} & {Learnable}  & {Sinusoidal} \\
        \midrule
        512 & \textbf{17.00} & 17.22 & 17.30 & 17.60 & 17.79 & 17.22 & 17.39 & 17.99 & 17.98 \\
        1024 & \textbf{16.52} & 16.73 & 16.79 & 17.11 & 17.26 & 16.70 & 16.89 & 17.47 & 17.48 \\
        2048 & \textbf{15.97} & 16.24 & 16.56 & 19.60 & 38.32 & 16.28 & 38.95 & \multicolumn{1}{c}{\textit{—}} & 219.80 \\
        4096 & \textbf{15.34} & 15.79 & 16.67 & 101.98 & 243.69 & 16.78 & 146.72 & \multicolumn{1}{c}{\textit{—}} & 1058.84 \\
        8192 & \textbf{15.41} & 15.97 & 17.23 & 383.08 & 799.53 & 20.32 & 361.26 & \multicolumn{1}{c}{\textit{—}} & 2485.84 \\
        15360 & \textbf{15.41} & 16.03 & 17.46 & 835.92 & 1450.83 & 26.13 & 691.90 & \multicolumn{1}{c}{\textit{—}} & 3355.86 \\
        \bottomrule

        \midrule
        \multicolumn{10}{c}{\textbf{WikiText-103}} \\
        \midrule 
        Sequence Length & {CABLE} & {{CABLE\textsubscript{NW}}} & {ALiBi} & {Fire} & {T5-bias} & {Kerple} & {RoPE} & {Learnable}  & {Sinusoidal} \\
        \midrule
        512 & 23.70 & 24.32 & 24.09 & 24.34 & 25.06 & 23.95 & \textbf{23.66} & 24.94 & 25.18 \\
        1024 & 22.32 & 23.01 & 22.74 & 22.90 & 23.60 & 22.56 & \textbf{22.26} & 23.53 & 23.73 \\
        2048 & \textbf{21.48} & 22.19 & 22.05 & 22.68 & 27.64 & 21.72 & 41.40 & \multicolumn{1}{c}{\textit{—}} & 172.33 \\
        4096 & \textbf{20.94} & 21.70 & 21.73 & 29.57 & 73.99 & 21.32 & 114.77 & \multicolumn{1}{c}{\textit{—}} & 607.48 \\
        8192 & \textbf{20.65} & 21.46 & 21.58 & 54.89 & 198.64 & 21.33 & 220.56 & \multicolumn{1}{c}{\textit{—}} & 1348.23 \\
        15360 & \textbf{20.33} & 21.13 & 21.30 & 104.79 & 411.09 & 21.58 & 375.62 & \multicolumn{1}{c}{\textit{—}} & 2017.42 \\
        \bottomrule
    \end{tabular}
    \caption{Perplexity comparison on the FineWeb-Edu-10B and WikiText-103 evaluation sets. The models in the upper table are GPT-2 Medium variants trained on the FineWeb-Edu-10B training set for 19k steps with a sequence length of 1024. The models in the lower table are GPT-2 Tiny variants trained on the WikiText-103 training set for 9k steps, also with a sequence length of 1024.}
    \label{tab:main_results}
\end{table*}

\textbf{Data-dependent Positional Encoding:} Recently, several data-dependent RPEs have been proposed. CoPE~\cite{golovneva2024contextual} applies fixed ALiBi-style biases with a learned binary gate. While effective in its domain (mathematics), it lacks the flexibility of continuous, learnable span control. DAPE~\cite{zheng2024dape, zheng2024dape2} functions as an augmentation to existing additive positional bias methods and is not a standalone mechanism. Furthermore, its architecture imposes a relatively high computational cost, as it uses feedforward or convolutional layers over the full attention matrix. FoX~\cite{lin2025forgetting}, a near-concurrent work, introduces forgetting mechanisms while minimizing reliance on positional encodings by learning token-wise biases.

Our method differs from these baselines, as CABLE avoids binarization and instead uses dynamic biases (unlike CoPE), functions as a standalone RPE and applies a lightweight MLP over the input (unlike DAPE), and additionally conditions the bias slope on the query token (unlike FoX).

\section{Proposed Method}

In this section, we formally introduce CABLE (\textbf{C}ontext-\textbf{A}ware \textbf{B}iases for \textbf{L}ength \textbf{E}xtrapolation), a novel additive relative positional encoding (RPE) approach designed to enhance the length generalization capabilities of Transformer models.

CABLE computes context-aware positional bias scores for each attention head and adds them to the pre-softmax attention logits. Unlike existing RPE methods, which are typically static and independent of the input sequence, our proposed biases are dynamically conditioned on the input context. Similar to the ALiBi method, we incorporate relative positional biases at the attention score level. However, CABLE introduces two key modifications: (1) the biases are learned and explicitly dependent on the input context, and (2) we learn distinct scalar weights for these biases. To implement this, we employ two separate linear layers—one to generate the context-aware biases and another to compute their associated weights.

Let $t$ and $d$ be the sequence length and the dimension of embeddings on each head, respectively. The learned bias for each token in the input sequence $X \in \mathbb{R}^{t \times d}$ is as follows:
\begin{align}
	\label{Eq:wc}
	f_\theta(X)=ReLU(Xw_c)
\end{align}
Where $f_\theta:\mathbb{R}^{t \times d} \xrightarrow{} \mathbb{R}_{+}^t$, and $\theta=\{w_c\ \in \mathbb{R}^d\}$.
Hence, we obtain context dependent biases for each token. Here ReLU  ensures the biases are positive. Then, by taking a cumlative sum of these biases  we inherently make these biases aware of how much positional bias should be incorporated until their position and obtain $S(X) \in \mathbb{R}_{+}^t$.
\begin{align}
    \label{Eq:cumsum}
    S(X) &= L f_\theta(X), \\
    L_{i,j} &=
    \begin{cases}
        1 & \text{if } j \leq i \\
        0 & \text{otherwise}
    \end{cases}
\end{align}
Where $L \in R^{t \times t}$ is a lower triangular matrix of ones.
Therefore, the relative biases $B(X) \in \mathbb{R}^{t \times t}$ for each pair of tokens in the input sequence is calculated by:
\begin{align}
    \label{Eq:rel_pos}
    B(X)_{i,j}=S(X)_i-S(X)_j
\end{align}
At this stage, we have obtained context-aware biases that can be directly added to the pre-softmax attention logits. In our experiments, we refer to this version—without any additional learnable parameters for the biases—as CABLE\textsubscript{NW}. To further enhance flexibility, we introduce a linear layer that learns a bias weight vector $g_\theta(X)$  for each token, allowing the model to modulate (dampen or amplify) the positional biases based on the input context.
\begin{align}
    \label{Eq:weights}
    g_\theta(X)=Softplus(Xw_s)
\end{align}
Where $g_\theta:\mathbb{R}^{t \times d} \xrightarrow{} \mathbb{R}_{+}^t$, and $\theta=\{w_s\ \in \mathbb{R}^d\}$.
Finally, we multiply each relative bias by its corresponding weight to produce the final CABLE bias for each attention head, as follows:
\begin{align}
    B(X)_{i,j}=g_\theta(X)_i (S(X)_i-S(X)_j)
\end{align}

CABLE exhibits an inductive bias similar to sliding window attention by penalizing distant query-key pairs—penalties that increase with positional distance. It can be seen as a generalization of ALiBi. while ALiBi applies fixed linear biases, CABLE learns context-aware biases for each token. Notably, if we set each token’s bias and weight to -1 and $1/2^h$ respectively, CABLE reduces to ALiBi, with relative biases simply reflecting token distances. However, CABLE’s key advantage is its ability to adapt these biases based on token context, enabling more expressive and flexible positional encoding.

As with most RPE methods, CABLE adds positional information only to the queries and keys (not the values), a practice shown to enhance length extrapolation in methods like ALiBi, T5-bias, and RoPE.

CABLE is simple, lightweight, and easily integrates into standard attention mechanisms. It requires only two additional linear layers, minimal parameters, and can be implemented in a few lines of code. The design involves two unfolding operations, a cumulative summation, and bias addition to the attention logits. Despite its simplicity, CABLE significantly improves extrapolation performance with negligible time and memory overhead compared to the vanilla transformer. Furthermore, it offers training time and memory usage on par with existing RPE methods, while maintaining low inference overhead and demonstrating notable gains in extrapolation, as shown in Section \ref{sec: results}.

\section{Experiment Setup}

\subsection{Datasets}
For training, we use the FineWeb dataset \cite{penedo2024fineweb}, a large-scale dataset (15 trillion tokens) for LLM pretraining, derived from 96 CommonCrawl snapshots. FineWeb has been shown to produce better-performing LLMs than other open pretraining datasets \cite{penedo2024fineweb}. More specifically, we use a 10B sample of the FineWeb-Edu dataset, which consists of 1.3T tokens from educational web pages filtered from the FineWeb dataset. We allocate 9.9B tokens for training and 0.1B for evaluation. Furthermore, we also train the models on WikiText-103 \cite{merity2016pointer}, a small dataset containing a preprocessed version of Wikipedia, widely used in many NLP tasks. For evaluation, we use the test sets of FineWeb-Edu, WikiText-103, and a 1B-token sample of the FineWeb dataset.

\subsection{Settings}
For all next-token prediction tasks, we use the GPT-2 variants \cite{brown2020language}. For the FineWeb-Edu-10B dataset, we use its small version (12 layers, 10 heads, and a hidden dimension of 768) with 124M parameters, and its medium version (24 layers, 16 heads, and a hidden dimension of 1024) with 334M parameters. We also incorporate a tiny version of GPT-2 (44M parameters) with 6 layers, 8 heads, and a hidden dimension of 512 for the WikiText-103 dataset, as it is a relatively small dataset. The evaluation metric is perplexity (PPL), and we train the models with sequence length of 1024. All the models are trained on eight H100 GPUs with 80G GPU RAM. Training settings are the same as those used for GPT-2 \cite{radford2019language}. Gradients are updated after processing 524,288 tokens and vocab size is 50304. For training on the FineWeb-Edu-10B dataset, we run 19k steps (\textasciitilde{}1 epoch) with batch sizes of 64, 32, and 16 for the tiny, small, and medium models, respectively. On WikiText-103, the tiny, small, and medium variants are trained for 9k, 5k, and 3k steps(\textasciitilde{}10, 5, 3 epochs respectively). The learning rate starts at 0.0006, with a linear warmup over 750 steps, followed by cosine decay to a minimum of 0.00006.

\begin{figure*}[!ht]
    \centering
    \begin{adjustbox}{max width=\linewidth}
        \includegraphics{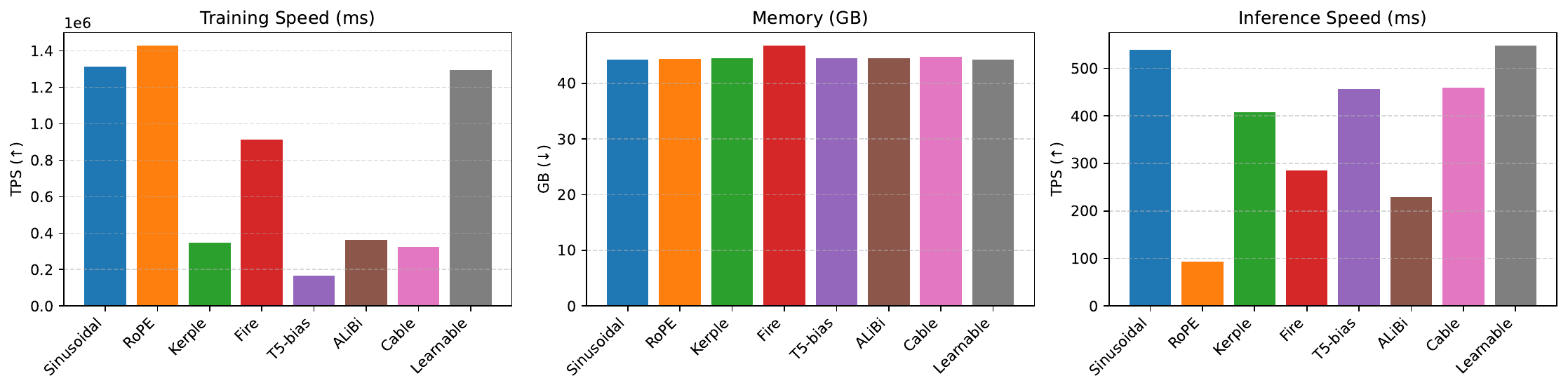}
    \end{adjustbox}
    \caption{Comparison of batched training time, memory usage in training, and unbatched inference time in GPT-2 Medium among the Sinusoidal, RoPE, Kerple, Fire, T5-bias, ALiBi, CABLE, and Learnable positional encoding methods.}
    \label{fig:time}
\end{figure*}

\subsection{Baselines}
We compare our method against the following positional encoding approaches:

\textbf{Learnable} \cite{vaswani2017attention}: A trainable APE where each position is associated with a learned embedding. The number of positions is fixed and predefined during training.

\textbf{Sinusoidal} \cite{vaswani2017attention}: A fixed APE used in early Transformer models \cite{vaswani2017attention, baevski2018adaptive, ott2018scaling, lewis2021base}.

\textbf{RoPE} \cite{su2024roformer}: A non-learnable non-additive RPE widely adopted in LLMs such as GPT-2 \cite{brown2020language}, LLaMA \cite{touvron2023llama}, PaLM \cite{chowdhery2023palm}, and Gemma \cite{team2024gemma, team2024gemma2}.

\textbf{ALiBi} \cite{press2021train}: A non-learnable additive RPE used in models like BLOOM \cite{le2023bloom} and Falcon \cite{almazrouei2023falcon}.

\textbf{T5-bias} \cite{raffel2020exploring}: A learnable additive RPE used in the T5 model.

\textbf{Kerple} \cite{chi2022kerple}: A learnable additive RPE with logarithmic and power variants; we use the logarithmic variant due to its superior performance.

\textbf{Fire} \cite{li2023functional}: A learnable additive RPE designed to give more weight to distant query-key pairs than other methods.

\section{Results}
\label{sec: results}
We evaluate the effectiveness of our method across multiple settings. First, we examine its extrapolation capability in decoder-only models for next-token prediction, comparing it against several representative RPEs as well as commonly used APEs. Next, we analyze runtime and memory efficiency, followed by a detailed comparison with another context-aware RPE, namely DAPE ~\cite{zheng2024dape2}. We then provide a visualization of the positional biases learned by CABLE. Finally, we conduct an ablation study, demonstrating that certain modifications can further enhance CABLE’s performance in specific scenarios.



\subsection{Length Extrapolation}
Our method demonstrates strong length extrapolation performance on the FineWeb-Edu-10B dataset, when trained with a sequence length of 1024 and evaluated on shorter and longer sequences, as shown in Figure \ref{fig:pull_figure}. Table \ref{tab:main_results} further compares the extrapolation capabilities of CABLE against baseline methods on the test sets of FineWeb-Edu-10B and WikiText-103. \footnote{For the longest sequences tested, we report results for 15,360 tokens instead of 16,384 due to computational constraints.}
The sinusoidal method suffers a sharp performance drop even with slight increases in sequence length. RoPE shows a similar trend—initial improvement followed by a significant decline at longer lengths. The learnable method performs competitively at 512 and 1024 tokens but lacks extrapolation ability beyond the training length.
T5-bias follows a similar trend, but its performance degrades more gradually than Sinusoidal. It initially extrapolates well to sequences slightly longer than those seen during training, thanks to its mechanism of learning relative positional information and reusing the maximum relative distance for unseen positions. ALiBi performs well on longer contexts overall but experiences slightly degradation at extreme lengths.

In contrast, our method consistently achieves lower PPL on longer sequences. Specifically, for models trained on sequence length of 1024, our method achieves lower PPL even when extrapolating to sequences 16 times longer. 

Moreover, on the FineWeb-Edu-10B dataset, which contains far more tokens than WikiText-103, a model trained with ALiBi on T=1024 performs well on T=2048 but begins to degrade with longer sequences. In contrast, CABLE shows consistent improvement, even for T=15360, and achieves a better PPL than it does on T=1024, the sequence length seen during training.

Our results demonstrate that the learned biases in CABLE capture contextual information more effectively than ALiBi, leading to superior length extrapolation, especially on very long sequences. Notably, even CABLE\textsubscript{NW} outperforms ALiBi, highlighting the strength of our context-aware design. Additionally, the improved performance with the full CABLE model underscores the benefit of the learned weight function $g_\theta(X)$.

\subsection{Runtime and Memory Overhead}
We also evaluate our method against existing methods in terms of training/inference runtime and memory usage. As shown in Figure \ref{fig:time}, our method achieves the same training Token Per Second (TPS) as ALiBi and Kerple. However, both our method and ALiBi have slightly higher overhead compared to RoPE, Sinusoidal, and Learnable methods, while T5-bias exhibits significant overhead. 
During inference, our method achieves faster performance than other RPEs. It is the third fastest overall—trailing only Sinusoidal and Learnable encodings—while outperforming ALiBi in speed.
Moreover, CABLE uses almost the same GPU memory as other methods during training and adds negligible overhead compared to methods like ALiBi. It should be noted that, due to the extrapolation ability of CABLE, it can be trained on shorter sequence lengths and effectively tested on much longer sequences. This approach addresses training overhead by reducing the sequence length during training, making it feasible on commonly available GPUs. The overhead of our method is primarily related to the cumulative sum operation in our computations. Importantly, for inference, we cache the cumulative sums, so there is no need to re-calculate them for all tokens each time. This optimization helps CABLE achieving superior inference time to other methods, such as ALiBi.


\begin{figure*}[!ht]
    \centering
    \begin{adjustbox}{max width=\linewidth}
        \includegraphics{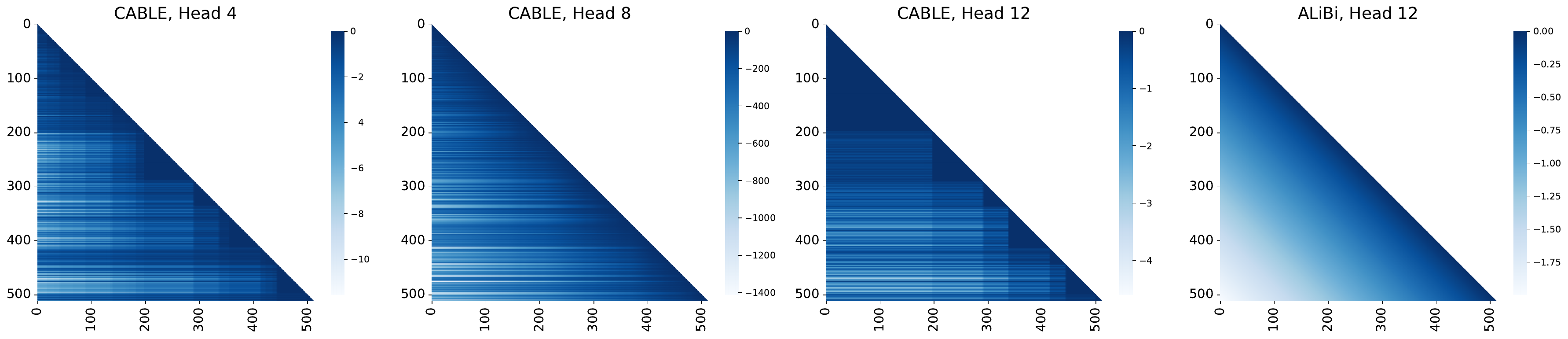}
    \end{adjustbox}
    \caption{Comparison of CABLE and ALiBi positional biases in GPT-2 Small. CABLE biases (left three panels) vary with context and attention head, while ALiBi biases (right panel) are fixed across all layers and tokens. This figure illustrates how CABLE can modulate attention selectively based on token context, capturing richer structural information compared to the uniform ALiBi biases.}
    \label{fig:cable_vs_alibi}
\end{figure*}

\subsection{Data-dependent RPEs: CABLE vs. DAPE}
As discussed in the related work section, a few context-aware RPEs have already been proposed, with DAPE~\cite{zheng2024dape} and its successor DAPEv2~\cite{zheng2024dape2} being the most notable. In this section, we compare CABLE against DAPEv2, since the latter represents a direct improvement over the former.

DAPE and DAPEv2 are data-dependent RPEs that act as augmentations to existing additive positional bias methods rather than standalone mechanisms. However, they come with considerable computational cost: DAPE relies on MLP layers, while DAPEv2 applies convolutional layers over the full attention matrix of shape $[B, nh, T, T]$, leading to higher training and backward-pass overhead. In contrast, CABLE applies a lightweight MLP over the input representations of shape $[B, T, D]$, resulting in significantly lower computational overhead.

Here we augment DAPEv2 with CABLE and Kerple for the comparison. Table~\ref{tab:dape} presents extrapolation results on the FineWeb-Edu-10B dataset, with models trained at sequence length 512 using GPT-2 Small. The results show that while DAPEv2 improves the performance of both CABLE and Kerple, the overall ranking remains unchanged: CABLE consistently outperforms Kerple, both with and without DAPEv2 augmentation. 

Kerple suffers from severe degradation at longer sequence lengths (from 4096 onward), whereas CABLE maintains robust performance even without augmentation. Applying DAPEv2 
to kerple can alleviate this weakness, but DAPEv2 introduces substantial computational overhead. For instance, GPT-2 Small with CABLE trains at ~1.9M tokens/sec, while DAPEv2+CABLE drops to ~0.6M tokens/sec (about 3× slower). Also at inference, DAPEv2+CABLE is ~1.7× slower than CABLE alone. Similar slowdowns are observed for Kerple with DAPEv2.

Overall, these findings suggest that CABLE offers the best trade-off between accuracy and efficiency, achieving state-of-the-art extrapolation quality while preserving high training and inference throughput.

\begin{table}[t]
\centering
\resizebox{\columnwidth}{!}{%
\begin{tabular}{c c c c c c}
\toprule
\textbf{Seq. Len.} & \textbf{CABLE} & \textbf{Kerple} & \textbf{DAPEv2+Cable} & \textbf{DAPEv2+Kerple} \\
\midrule
512  & 21.17 & 21.41  & \textbf{20.41} & 20.56 \\
1024 & 20.72 & 21.12  & \textbf{19.91} & 20.07 \\
2048 & 20.23 & 22.58  & \textbf{19.30} & 19.51 \\
4096 & 19.60 & 28.04  & \textbf{18.55} & 18.79 \\
8192 & 19.87 & 39.38  & \textbf{18.63} & 18.92 \\
\bottomrule
\end{tabular}}
\caption{Extrapolation results for CABLE and Kerple when augmented with DAPEv2.}
\label{tab:dape}
\end{table}

\subsection{Analysis of Positional Biases}
In this section, we analyze the positional biases added by CABLE and ALiBi to the original attention scores, with a focus on how contextual information influences these biases. Figure \ref{fig:cable_vs_alibi} illustrates the relative biases for three different heads in the last layer, alongside the ALiBi biases. Unlike ALiBi, which applies a fixed bias uniformly across all layers and tokens, CABLE biases are both head- and context-dependent. Appendix \ref{app:a3} presents the complete set of visualizations for all attention heads across all examined methods.

The biases presented in Figure \ref{fig:cable_vs_alibi} were computed using a GPT-2 Small model trained and evaluated with a sequence length of 512 tokens. The contextual dependence of CABLE can be interpreted as a mechanism that allows the model to dynamically calibrate positional information based on the structure of the input sequence. For example, tokens that serve as anchors (e.g., punctuation, repeated entities, or syntactic markers) may receive weaker biases, enabling the model to maintain stronger attention to them even at long distances. In contrast, less informative tokens may be more heavily penalized. Such adaptive behavior could explain why CABLE provides stronger representational flexibility and, in practice, has been observed to improve extrapolation and performance on tasks requiring nuanced handling of context length. 
Taken together, these findings suggest that while ALiBi provides a simple and efficient positional prior, its uniformity across layers and tokens constrains its ability to model diverse contextual dependencies. By contrast, CABLE introduces a more expressive bias structure that adapts to both token content and attention head, potentially offering a more faithful integration of positional and semantic information.

\subsection{Bidirectional Models}
In this major experiment, we evaluate the effectiveness of our proposed additive RPE method in learning contextual representations. To do this, we replace the original fixed learnable positional encodings in BERT \cite{devlin-etal-2019-bert} with CABLE during pre-training. We use the 10B-sample FineWeb-Edu dataset and train the models using only the masked language modeling (MLM) objective, following \citet{liu2019roberta}, who showed that removing next sentence prediction (NSP) can improve performance. Our BERT models are based on the bert-base-uncased architecture and are trained on four H100 GPUs with a batch size of 32 and a maximum sequence length of 512 for 14k steps (\textasciitilde{}1 epoch on FineWeb-Edu-10B). We use Adam \cite{kingma2014adam} with a learning rate of 1e-4. As shown in Figure~\ref{sec:appendix}, CABLE achieves faster convergence in MLM loss compared to other positional encoding baselines.

Since RPE methods do not impose strict limitations on context length, BERT models trained with these encodings show improved long-context performance compared to existing encoder-only models. However, standardized long-context benchmarks for encoder-only architectures remain limited. Following \citet{warner2024smarter}, we evaluate long-context performance using the English subset of MLDR \cite{chen2024bge}, a retrieval benchmark consisting of over 200,000 long documents.

To adapt BERT models for this task, we fine-tune them on MS-MARCO \cite{nguyen2016ms} using mined hard negatives \cite{xuan2020hard}, with 1.25M samples, a batch size of 128, and a 5\% learning rate warmup over one epoch, leveraging the sentence-transformers framework \cite{reimers2019sentence}. We then evaluate the fine-tuned models on the MLDR test set using nDCG@10 as the evaluation metric. Table~\ref{tab:MLDR} presents the results for several competitive positional encoding methods.\footnote{We report only ALiBi among additive RPEs for BERT, as it outperformed other variants in our experiments.}

At shorter sequence lengths (512 tokens), RoPE performs slightly better than other methods. However, as sequence length increases, CABLE consistently outperforms all baselines, showing notable gains beyond 1024 tokens. ALiBi maintains reasonable performance but still trails behind CABLE. In contrast, RoPE’s effectiveness drops sharply after 1024 tokens, becoming nearly unusable at longer lengths. Learnable absolute encodings perform not competitively even at 512 tokens and cannot generalize to longer sequences. Sinusoidal encoding is effective at short lengths but fails completely beyond 1024 tokens. Overall, CABLE exhibits the strongest scalability to long sequences, while others either degrade significantly or become unusable.

\begin{table}[t]
    \centering
    \footnotesize
    \begin{adjustbox}{max width=\linewidth}
    \begin{tabular}{l 
                    S[table-format=2.2]
                    S[table-format=2.2]
                    S[table-format=2.2]
                    S[table-format=2.2]
                    S[table-format=2.2]}
        \toprule
        \multicolumn{6}{c}{\textbf{MLDR – nDCG@10 vs. Sequence Length}} \\
        \midrule
        Seq. Len. 
        & {CABLE}
        & {ALiBi} 
        & {RoPE} 
        & {Learnable} 
        & {Sinusoidal} \\
        \midrule
        512   & 14.96 & 14.02 & \textbf{15.16} & 10.42 & 13.57 \\
        1024  & \textbf{15.15} & 12.88 & 14.41 & \multicolumn{1}{c}{\textit{—}} & 12.80 \\
        2048  & \textbf{16.77} & 14.30 & 10.26 & \multicolumn{1}{c}{\textit{—}} & 1.03 \\
        4096  & \textbf{21.36} & 18.71 & 1.17  & \multicolumn{1}{c}{\textit{—}} & 0.00 \\
        8192  & \textbf{24.59} & 22.86 & 0.12  & \multicolumn{1}{c}{\textit{—}} & 0.00 \\
        16384 & \textbf{25.10} & 23.44 & 0.12  & \multicolumn{1}{c}{\textit{—}} & 0.00 \\
        \bottomrule
    \end{tabular}
    \end{adjustbox}
    \caption{Retrieval performance (nDCG@10) on the MLDR test set for BERT models with different positional encodings, trained at sequence length 512 and evaluated on longer inputs.}
     \label{tab:MLDR}
\end{table}

\subsection{Ablation Study: Kernelized CABLE}

\begin{figure}[t]
    \vspace{-0.5em} 
    \hfill
     \begin{adjustbox}{max width=\columnwidth}
        \centering
        \includegraphics[width=\linewidth]{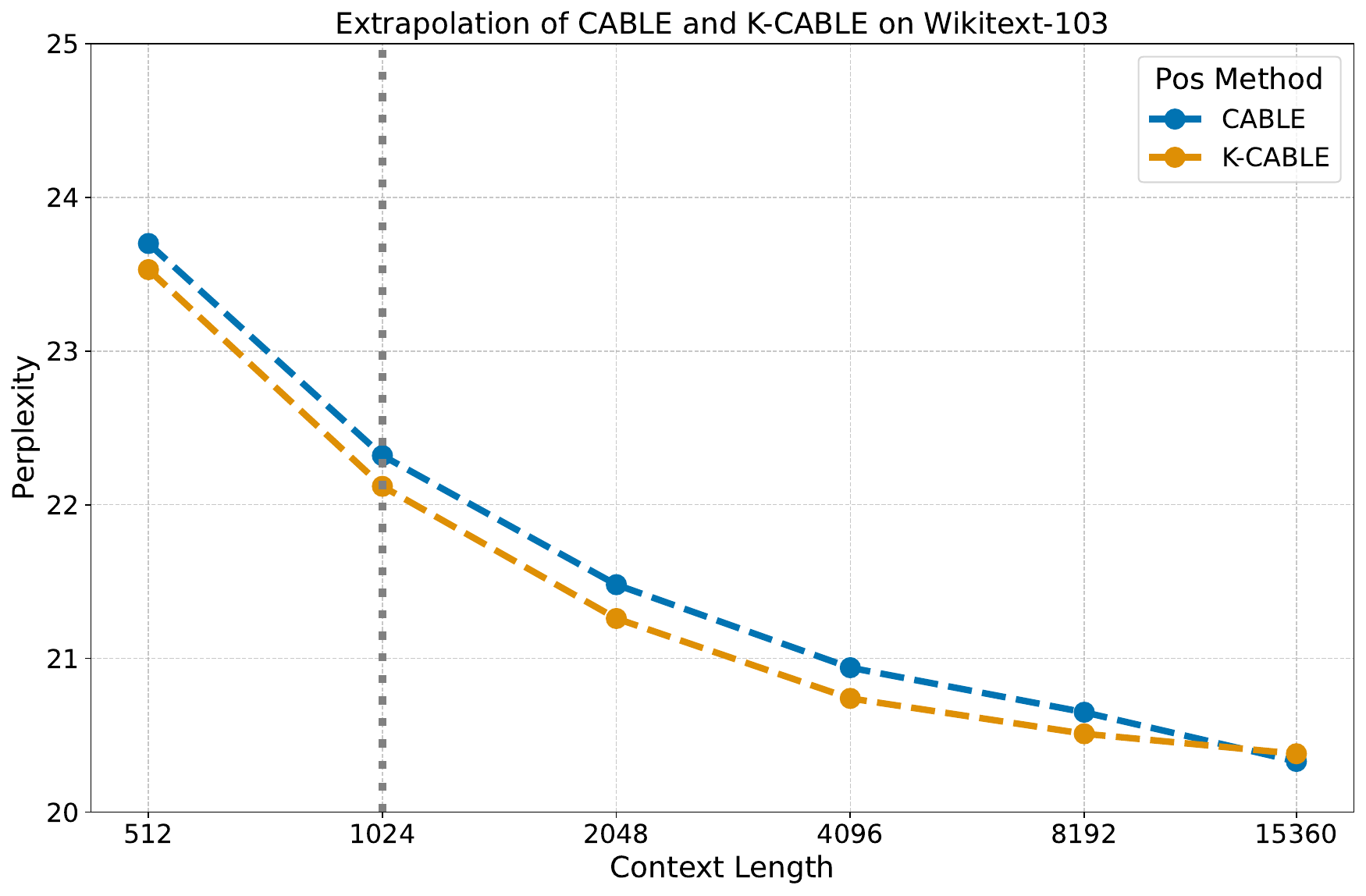}
    \end{adjustbox}
        \caption{Extrapolation performance of CABLE vs. its kernelized variant (K-CABLE). Both models are GPT-2 Tiny architectures trained on the WikiText-103 training set with a sequence length of 1024.}
        \label{fig:kcable}
\end{figure}

For models with a low number of layers, we tested a kernelized version of our method (K-CABLE). In this setting, we use $-\log{(b^2+1)}$ as a kernel, which is applied to the relative biases before adding them to the attention scores. Figure \ref{fig:kcable} shows the extrapolation comparison between CABLE and K-CABLE. As can be seen, K-CABLE achieves better PPL when trained on sequence length of 1024, compared to original CABLE, demonstrating improved extrapolation. 
This improvement is related to the sliding window nature of additive RPEs, where, with a low number of layers, they struggle to propagate information across longer sequences. In contrast, K-CABLE has a slower slope for biases and behaves less like a sliding window, making it more suitable for networks with fewer layers. More generally, different types of kernels can be applied to CABLE based on the network architecture, allowing it to achieve optimal performance.

\section{Conclusion}
We introduced CABLE, a novel additive relative positional encoding method that learns context-aware biases for each token by injecting them into the attention matrix at every decoder layer. Unlike constant linear biases in ALiBi, CABLE adapts to tokens’ roles within the sequence. Experiments show that CABLE lowers perplexity, significantly improves length extrapolation, and consistently outperforms baselines on edu-fineweb10B and wikitext-103. Moreover, CABLE improves the long-context retrieval performance of encoder-only models.
These gains come with only minimal training overhead but faster inference compared to the existing RPEs.

\section*{Limitation}
While CABLE demonstrates strong performance in length extrapolation, it has several limitations. First, it incurs higher training time compared to RoPE due to its dynamic bias computation, though this overhead is negligible in inference. Second, CABLE occasionally underperforms RoPE at base sequence lengths (e.g., 1024 tokens in our experiments), particularly in tasks where fixed positional patterns suffice, suggesting a trade-off between adaptability and consistency for shorter contexts. Additionally, the method’s computational overhead, though minimal, may become more pronounced for extremely long sequences (>100K tokens), and its extrapolation capabilities remain dependent on the diversity of positional patterns in training data. While empirical results are promising, theoretical analysis of its attention dynamics at arbitrary lengths remains an open question. Future work could explore optimizations for training efficiency and head-specific bias adaptation to further enhance flexibility.

\bibliography{main}
\bibliographystyle{acl_natbib}

\newpage
\appendix
\section{Appendix}
\label{sec:appendix}

\subsection{Bert Models Training}
\begin{figure}[!ht]
    \centering
    \begin{adjustbox}{max width=\columnwidth}
        \includegraphics{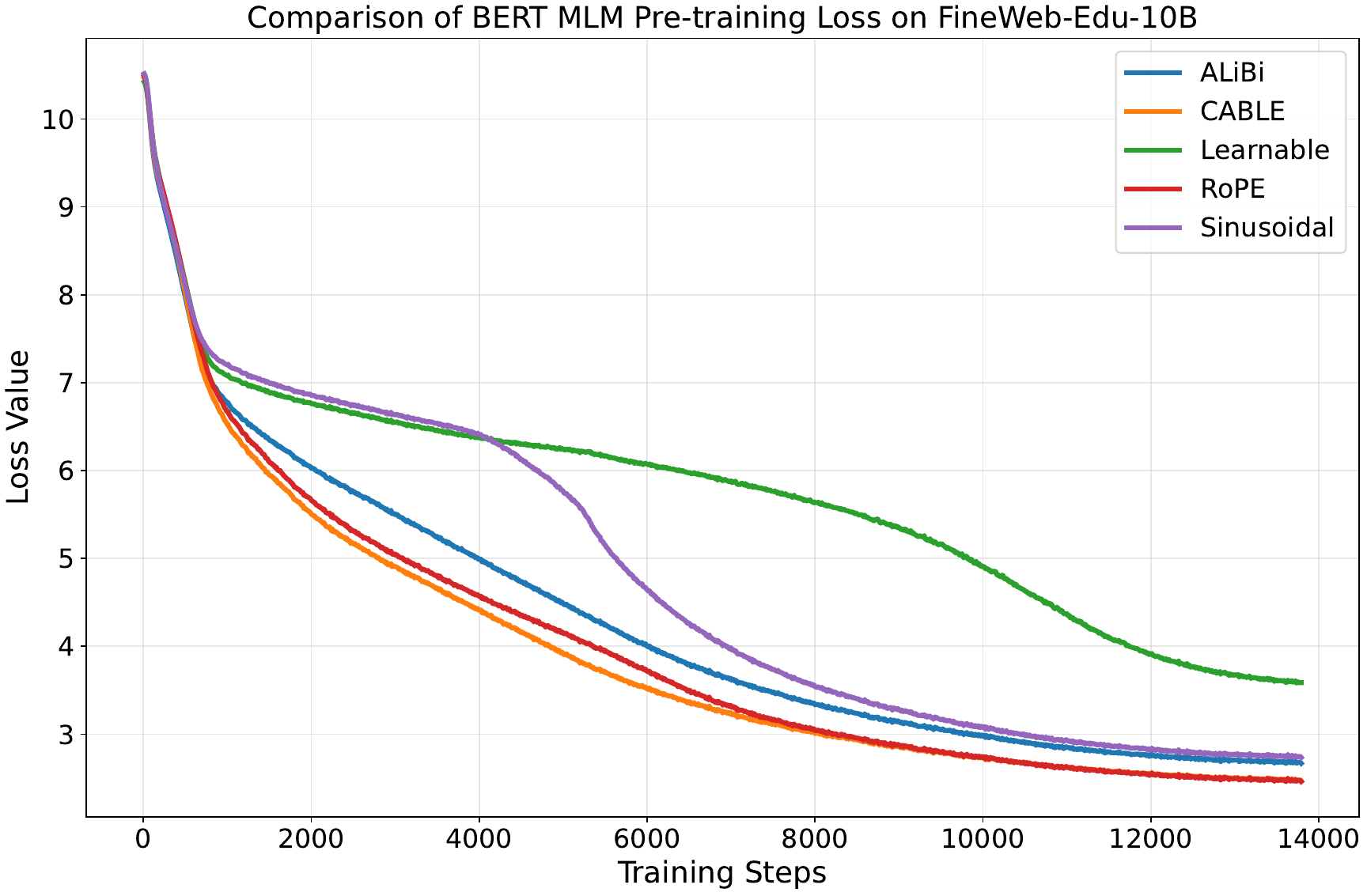}
    \end{adjustbox}
    \caption{Masked language modeling (MLM) loss during BERT pre-training on FineWeb-Edu-10B with different positional encoding methods. CABLE achieves the fastest convergence and lowest final loss, demonstrating superior training efficiency over traditional and other RPE methods.}
    \label{fig:bert_pre}
\end{figure}

Figure \ref{fig:bert_pre} shows the masked language modeling loss during BERT pre-training with different positional encodings. Traditional methods like learnable and sinusoidal fail to match the loss achieved by RPEs, highlighting the effectiveness of RPEs. CABLE also converges faster than other methods.

\begin{figure}[!ht]
    \centering
    \begin{adjustbox}{max width=\columnwidth}
        \includegraphics{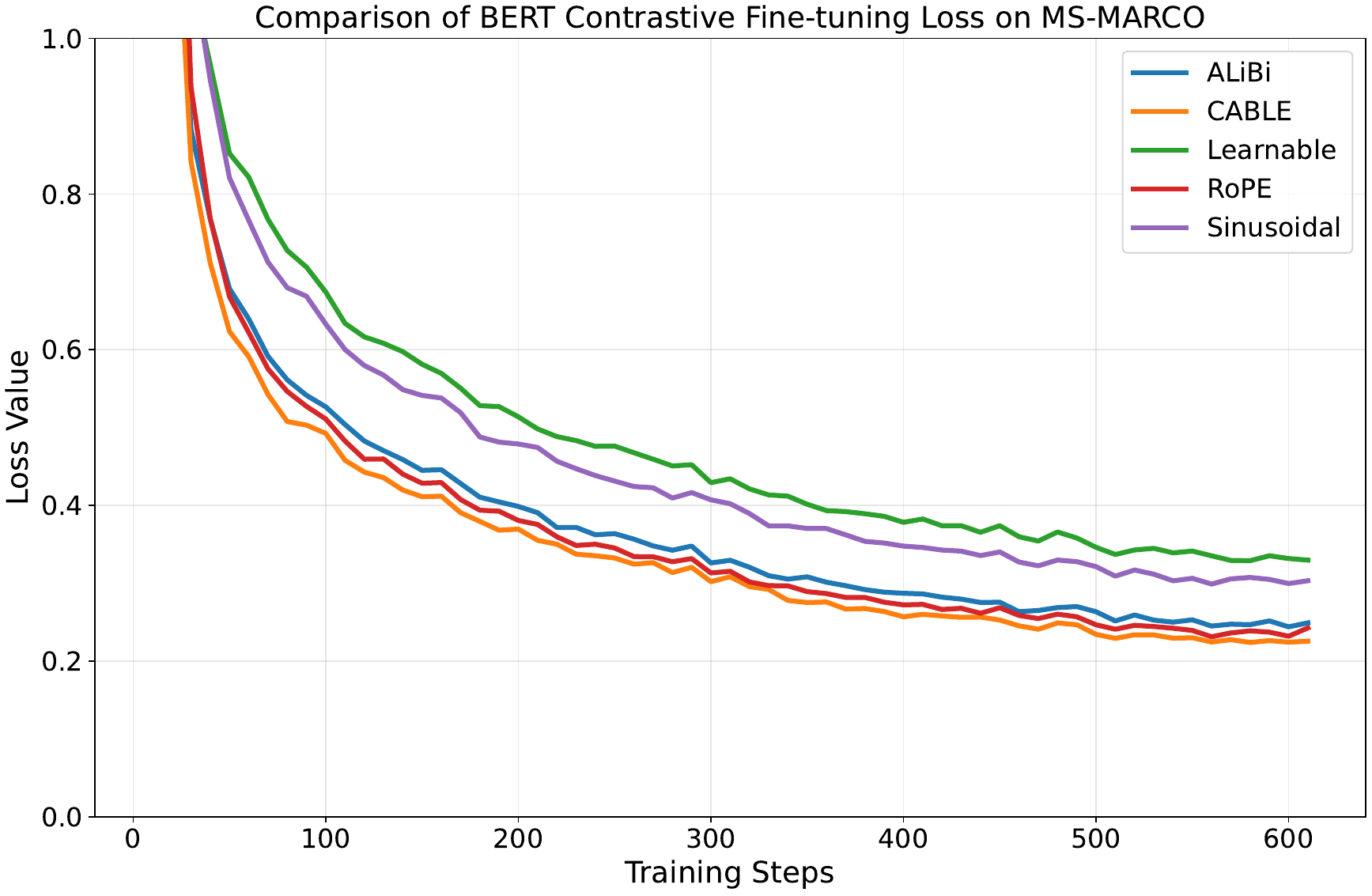}
    \end{adjustbox}
    \caption{Contrastive fine-tuning loss on the MS-MARCO dataset for BERT models using different positional encoding methods. CABLE achieves the lowest loss, while learnable and sinusoidal encodings underperform.}
    \label{fig:bert_fine}
\end{figure}

Figure \ref{fig:bert_fine} shows the contrastive loss during fine-tuning BERT models with different positional encoding methods on the MS-MARCO training set. Once again, learnable and sinusoidal methods lag behind, while CABLE achieves the lowest loss among all methods.

\subsection{Results for GPT-2 Small and Tiny}
Tables~\ref{tab:main_results2} and~\ref{tab:main_results3} present the extrapolation results for GPT-2 Small and GPT-2 Tiny models evaluated on the FineWeb-Edu-10B dataset. As shown in Table~\ref{tab:main_results2}, CABLE consistently outperforms other positional encoding methods across all sequence lengths, particularly at extrapolated lengths beyond 1024 tokens. Both GPT-2 Small and GPT-2 Tiny models trained with CABLE achieve significantly lower perplexity than those using ALiBi, RoPE, T5-bias, and other baselines. Notably, standard methods such as sinusoidal or learnable encodings degrade sharply at longer lengths, whereas CABLE maintains stable and superior performance. These results further confirm CABLE's effectiveness in enhancing length extrapolation, even in smaller model regimes.

\begin{table*}
    \centering
    \footnotesize
    \setlength{\tabcolsep}{3pt}
    \begin{tabular}{l S[table-format=2.2] S[table-format=2.2] *{9}{c}}
        \toprule
        \midrule
        \multicolumn{10}{c}{\textbf{GPT-2 Small}} \\
        \midrule 
        Sequence Length & {CABLE} & {{CABLE\textsubscript{NW}}} & {ALiBi} & {Fire} & {T5-bias} & {Kerple} & {RoPE} & {Learnable}  & {Sinusoidal} \\
        \midrule
        512 & \textbf{21.19} & 21.42 & 21.55 & 21.84 & 22.17 & 21.46 & 21.43 & 22.16 & 22.38 \\
        1024 & \textbf{20.63} & 20.89 & 20.99 & 21.26 & 21.57 & 20.86 & 20.87 & 21.56 & 21.83 \\
        2048 & \textbf{20.02} & 20.34 & 20.67 & 22.48 & 29.37 & 20.38 & 58.59 & - & 207.53 \\
        4096 & \textbf{19.24} & 19.67 & 21.23 & 53.25 & 131.36 & 21.11 & 225.78 & - & 956.41 \\
        8192 & \textbf{19.31} & 19.81 & 22.42 & 155.32 & 405.94 & 26.59 & 554.12 & - & 2376.51 \\
        15360 & \textbf{19.28} & 19.82 & 22.89 & 333.91 & 757.36 & 34.91 & 957.87 & - & 3589.97 \\
        \bottomrule

        \midrule
        \multicolumn{10}{c}{\textbf{GPT-2 Tiny}} \\
        \midrule 
        Sequence Length & {CABLE} & {{CABLE\textsubscript{NW}}} & {ALiBi} & {Fire} & {T5-bias} & {Kerple} & {RoPE} & {Learnable}  & {Sinusoidal} \\
        \midrule
        512 & \textbf{29.37} & 30.12 & 29.88 & 30.23 & 30.78 & 29.60 & 29.44 & 30.73 & 30.67 \\
        1024 & \textbf{28.73} & 29.57 & 29.25 & 29.56 & 30.08 & 28.95 & 28.81 & 30.11 & 30.03 \\
        2048 & \textbf{27.96} & 28.88 & 28.82 & 29.60 & 33.81 & 28.32 & 76.29 & \multicolumn{1}{c}{\textit{—}} & 275.28 \\
        4096 & \textbf{26.90} & 27.85 & 28.28 & 37.86 & 86.33 & 28.31 & 239.95 & \multicolumn{1}{c}{\textit{—}} & 1166.46 \\
        8192 & \textbf{26.97} & 27.92 & 26.80 & 70.72 & 222.60 & 32.00 & 452.67 & \multicolumn{1}{c}{\textit{—}} & 2561.54 \\
        15360 & \textbf{26.80} & 27.75 & 28.52 & 124.29 & 448.08 & 37.67 & 652.52 & \multicolumn{1}{c}{\textit{—}} & 3679.78 \\
        \bottomrule
    \end{tabular}
    \caption{Perplexity comparison on the FineWeb-Edu-10B evaluation sets. The upper table shows GPT-2 Small variants, and the lower table shows GPT-Tiny variants—both trained on the FineWeb-Edu-10B training set for 19k steps with a sequence length of 1024.}
    \label{tab:main_results2}
\end{table*}

\begin{table*}
    \centering
    \footnotesize
    \setlength{\tabcolsep}{3pt}
    \begin{tabular}{l S[table-format=2.2] S[table-format=2.2] *{9}{c}}
        \toprule
        \midrule
        \multicolumn{10}{c}{\textbf{GPT-2 Medium}} \\
        \midrule 
        Sequence Length & {CABLE} & {{CABLE\textsubscript{NW}}} & {ALiBi} & {Fire} & {T5-bias} & {Kerple} & {RoPE} & {Learnable}  & {Sinusoidal} \\
        \midrule
        512 & \textbf{20.33} & 20.80 & 20.80 & 22.18 & 23.33 & 20.96 & 20.81 & 22.52 & 24.16 \\
        1024 & \textbf{19.12} & 19.63 & 19.62 & 20.89 & 22.00 & 19.73 & 19.60 & 21.23 & 22.77 \\
        2048 & \textbf{18.36} & 18.91 & 19.04 & 21.86 & 35.46 & 19.01 & 20.78 & \multicolumn{1}{c}{\textit{—}} & 143.58 \\
        4096 & \textbf{17.87} & 18.47 & 18.91 & 46.60 & 124.19 & 18.63 & 31.22 & \multicolumn{1}{c}{\textit{—}} & 467.51 \\
        8192 & \textbf{17.58} & 18.23 & 18.89 & 106.27 & 363.59 & 18.61 & 51.54 & \multicolumn{1}{c}{\textit{—}} & 1006.37 \\
        15360 & \textbf{17.36} & 17.95 & 18.60 & 195.93 & 726.18 & 18.79 & 91.53 & \multicolumn{1}{c}{\textit{—}} & 1516.97 \\
        \bottomrule

        \midrule
        \multicolumn{10}{c}{\textbf{GPT-2 Small}} \\
        \midrule 
        Sequence Length & {CABLE} & {{CABLE\textsubscript{NW}}} & {ALiBi} & {Fire} & {T5-bias} & {Kerple} & {RoPE} & {Learnable}  & {Sinusoidal} \\
        \midrule
        512 & \textbf{20.93} & 21.34 & 21.46 & 22.03 & 22.80 & 21.50 & 21.38 & 22.51 & 23.09 \\
        1024 & \textbf{19.71} & 20.15 & 20.22 & 20.74 & 21.49 & 20.22 & 20.13 & 21.21 & 21.73 \\
        2048 & \textbf{18.95} & 19.42 & 19.62 & 21.28 & 33.85 & 19.45 & 30.14 & \multicolumn{1}{c}{\textit{—}} & 163.49 \\
        4096 & \textbf{18.47} & 18.98 & 19.39 & 36.02 & 123.05 & 19.04 & 63.75 & \multicolumn{1}{c}{\textit{—}} & 580.86 \\
        8192 & \textbf{18.20} & 18.75 & 19.29 & 81.77 & 347.22 & 18.91 & 117.81 & \multicolumn{1}{c}{\textit{—}} & 1121.08 \\
        15360 & \textbf{17.92} & 18.48 & 19.06 & 163.91 & 659.72 & 19.00 & 202.23 & \multicolumn{1}{c}{\textit{—}} & 1652.53 \\
        \bottomrule
    \end{tabular}
    \caption{Perplexity comparison on the WikiText-103 evaluation set. The upper table shows GPT-2 Medium variants trained for 3k steps, and the lower table shows GPT-2 Small variants trained for 5k steps. All models use a sequence length of 1024.}
    \label{tab:main_results3}
\end{table*}

\subsection{Visualizations}
\label{app:a3}
\begin{figure*}[!ht]
    \centering
    \begin{adjustbox}{max width=\linewidth}
        \includegraphics{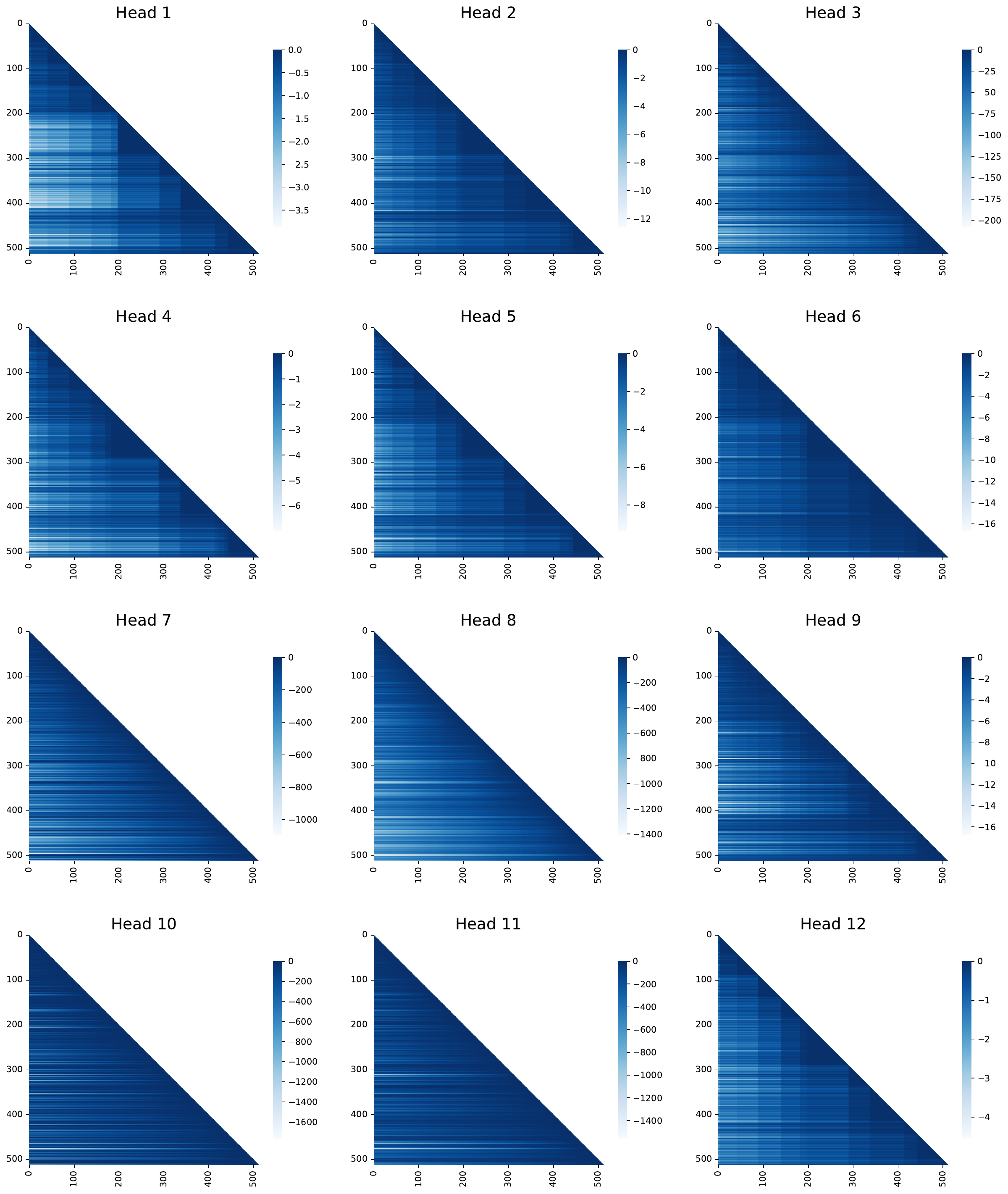}
    \end{adjustbox}
    \caption{Visualization of \textbf{CABLE} positional biases across all attention heads in the final layer of GPT-2 Small, shown for a randomly sampled input sequence. As like other additive RPEs, these biases are added to the attention scores. The varying patterns across tokens highlight the context-dependent nature of CABLE.}
    \label{fig:all_heads_cable}
\end{figure*}

\begin{figure*}[!ht]
    \centering
    \begin{adjustbox}{max width=\linewidth}
        \includegraphics{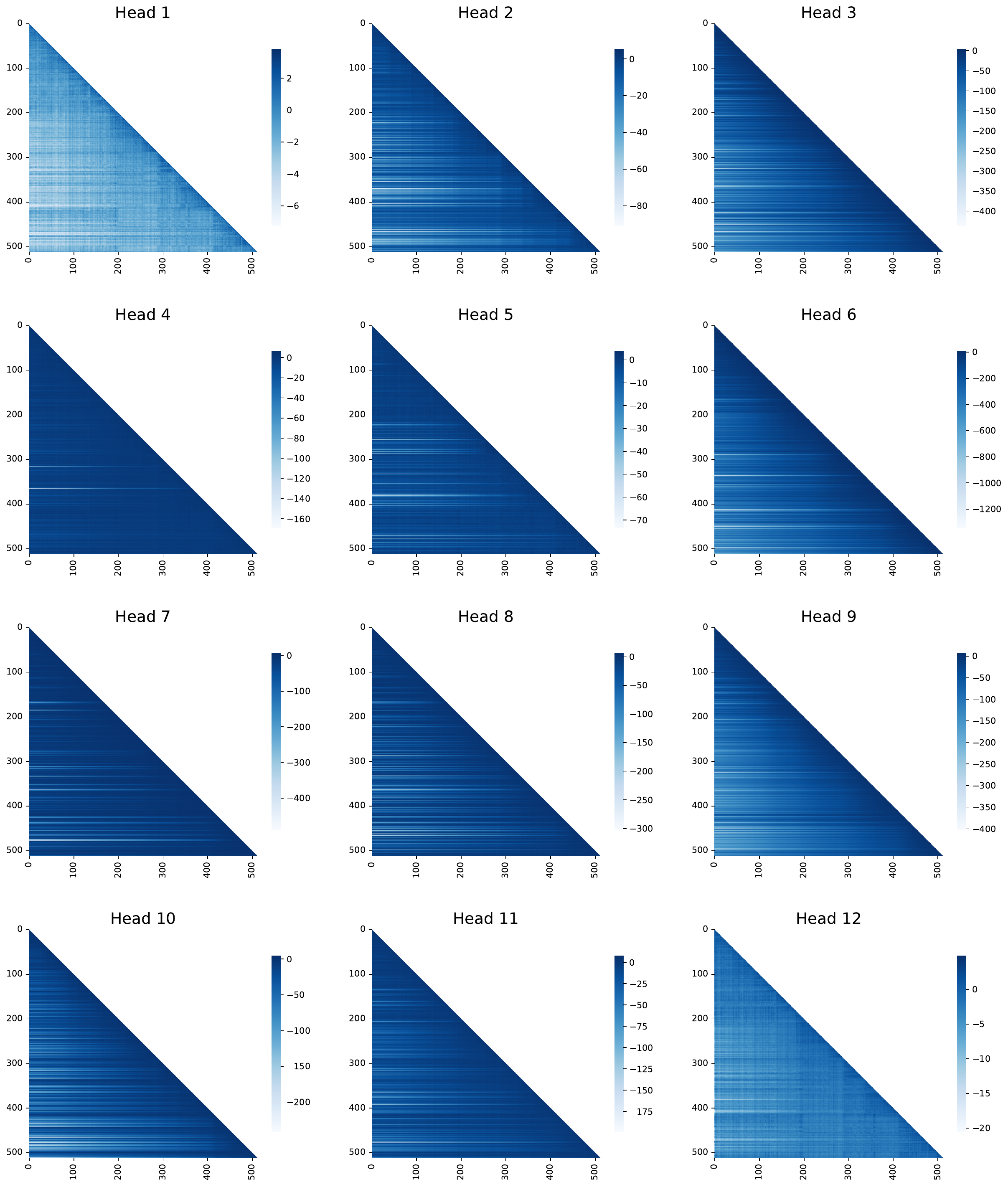}
    \end{adjustbox}
    \caption{Visualization of \textbf{DAPEv2+CABLE} positional biases across all attention heads in the final layer of GPT-2 Small, shown for a randomly sampled input sequence. As like other additive RPEs, these biases are added to the attention scores. The varying patterns across tokens highlight the context-dependent nature of CABLE.}
    \label{fig:all_heads_cabledape2}
\end{figure*}

\begin{figure*}[!ht]
    \centering
    \begin{adjustbox}{max width=\linewidth}
        \includegraphics{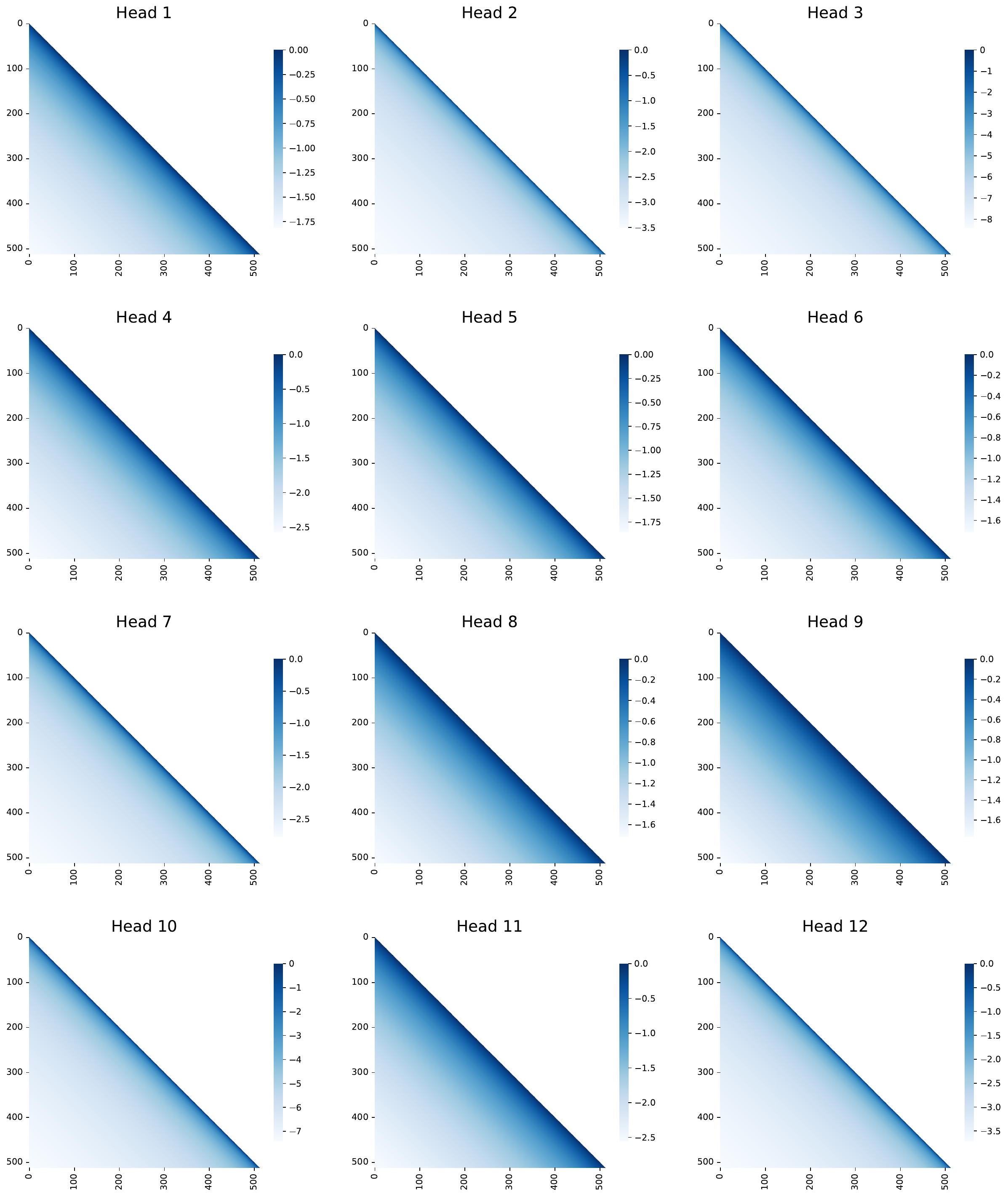}
    \end{adjustbox}
    \caption{Visualization of \textbf{Kerple} positional biases across all attention heads in the final layer of GPT-2 Small, shown for a randomly sampled input sequence. As like other additive RPEs, these biases are added to the attention scores. The uniform patterns across tokens highlight the context-independent nature of Kerple.}
    \label{fig:all_heads_kerple}
\end{figure*}

\begin{figure*}[!ht]
    \centering
    \begin{adjustbox}{max width=\linewidth}
        \includegraphics{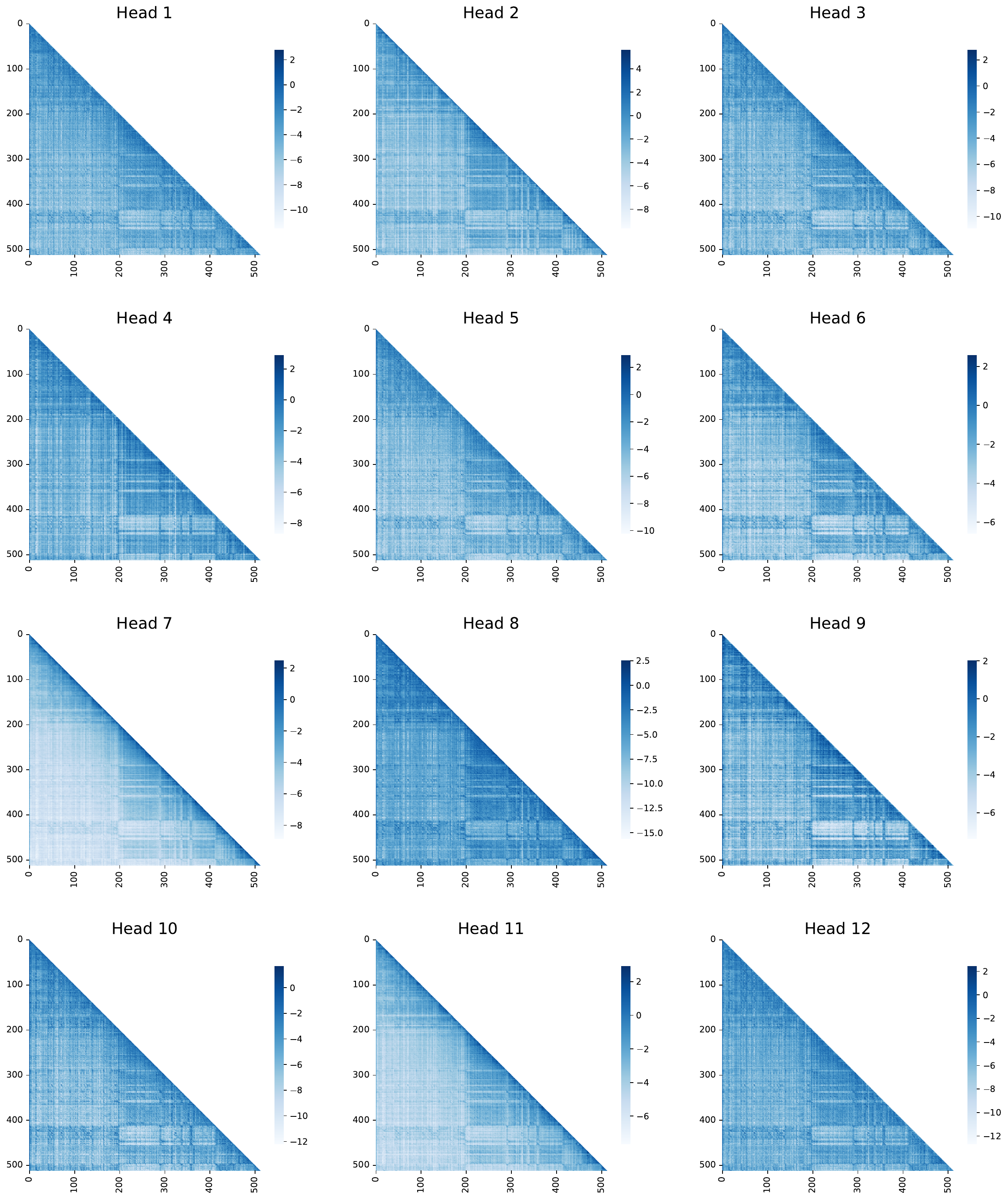}
    \end{adjustbox}
    \caption{Visualization of \textbf{DAPEv2+Kerple} positional biases across all attention heads in the final layer of GPT-2 Small, shown for a randomly sampled input sequence. As like other additive RPEs, these biases are added to the attention scores. The varying patterns across tokens highlight the context-dependent nature of Kerple when augmented with DAPEv2.}
    \label{fig:all_heads_kerpledape2}
\end{figure*}

\begin{figure*}[!ht]
    \centering
    \begin{adjustbox}{max width=\linewidth}
        \includegraphics{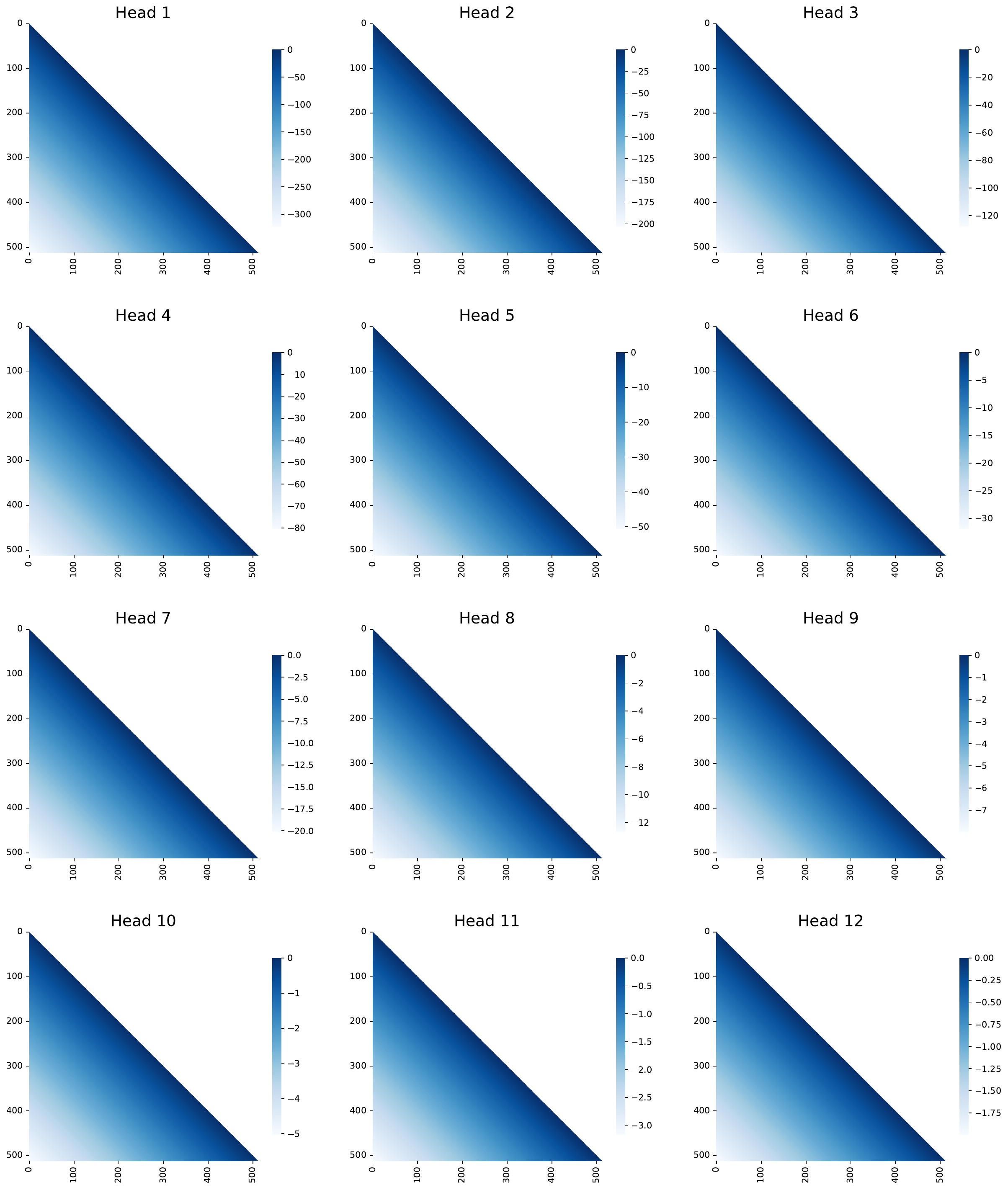}
    \end{adjustbox}
    \caption{Visualization of \textbf{ALiBi} positional biases across all attention heads in the final layer of GPT-2 Small, shown for a randomly sampled input sequence. As like other additive RPEs, these biases are added to the attention scores. The uniform patterns across tokens highlight the context-independent nature of ALiBi.}
    \label{fig:all_heads_alibi}
\end{figure*}

\begin{figure*}[!ht]
    \centering
    \begin{adjustbox}{max width=\linewidth}
        \includegraphics{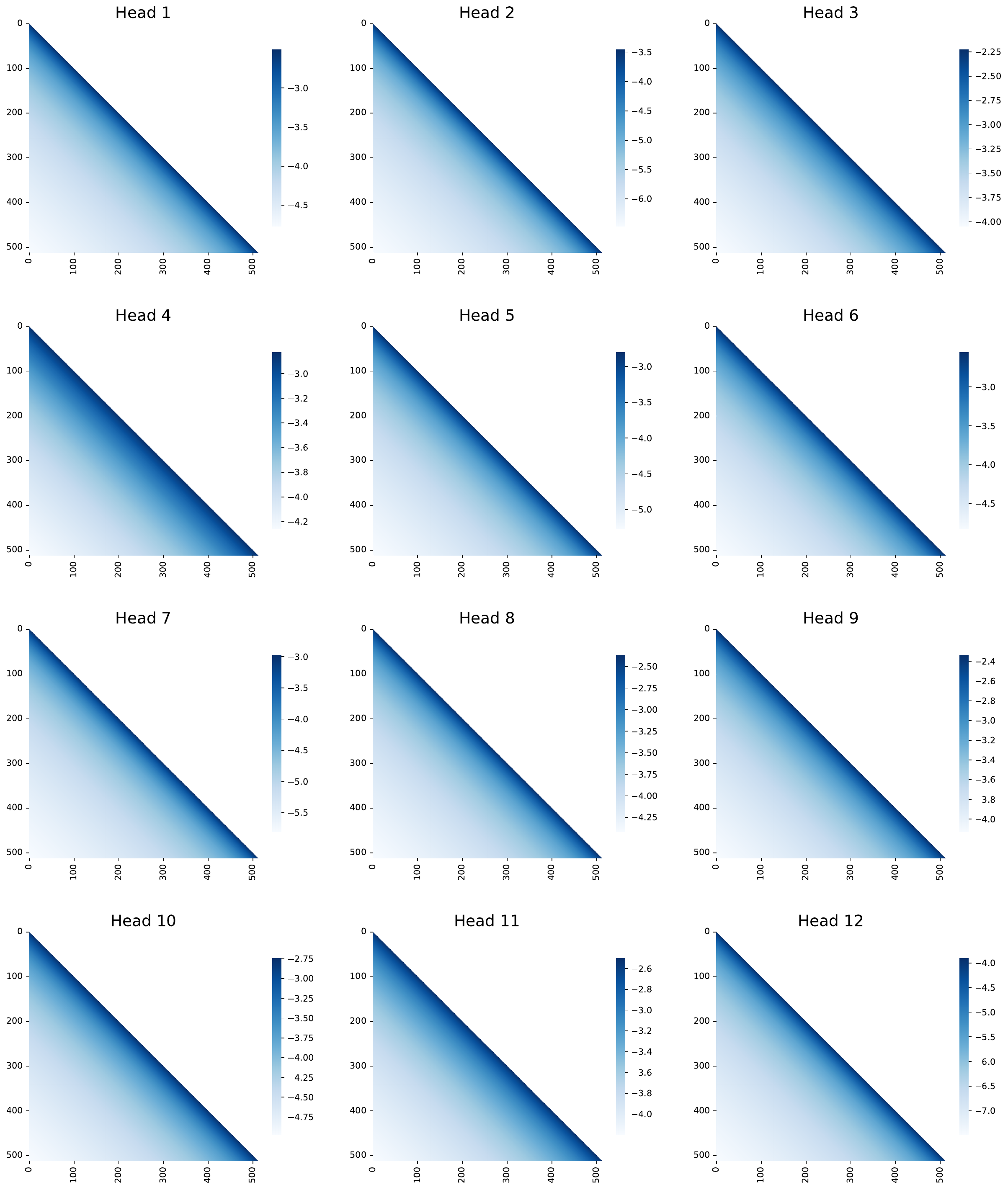}
    \end{adjustbox}
    \caption{Visualization of \textbf{Fire} positional biases across all attention heads in the final layer of GPT-2 Small, shown for a randomly sampled input sequence. As like other additive RPEs, these biases are added to the attention scores. The uniform patterns across tokens highlight the context-independent nature of Fire.}
    \label{fig:all_heads_fire}
\end{figure*}

\begin{figure*}[!ht]
    \centering
    \begin{adjustbox}{max width=\linewidth}
        \includegraphics{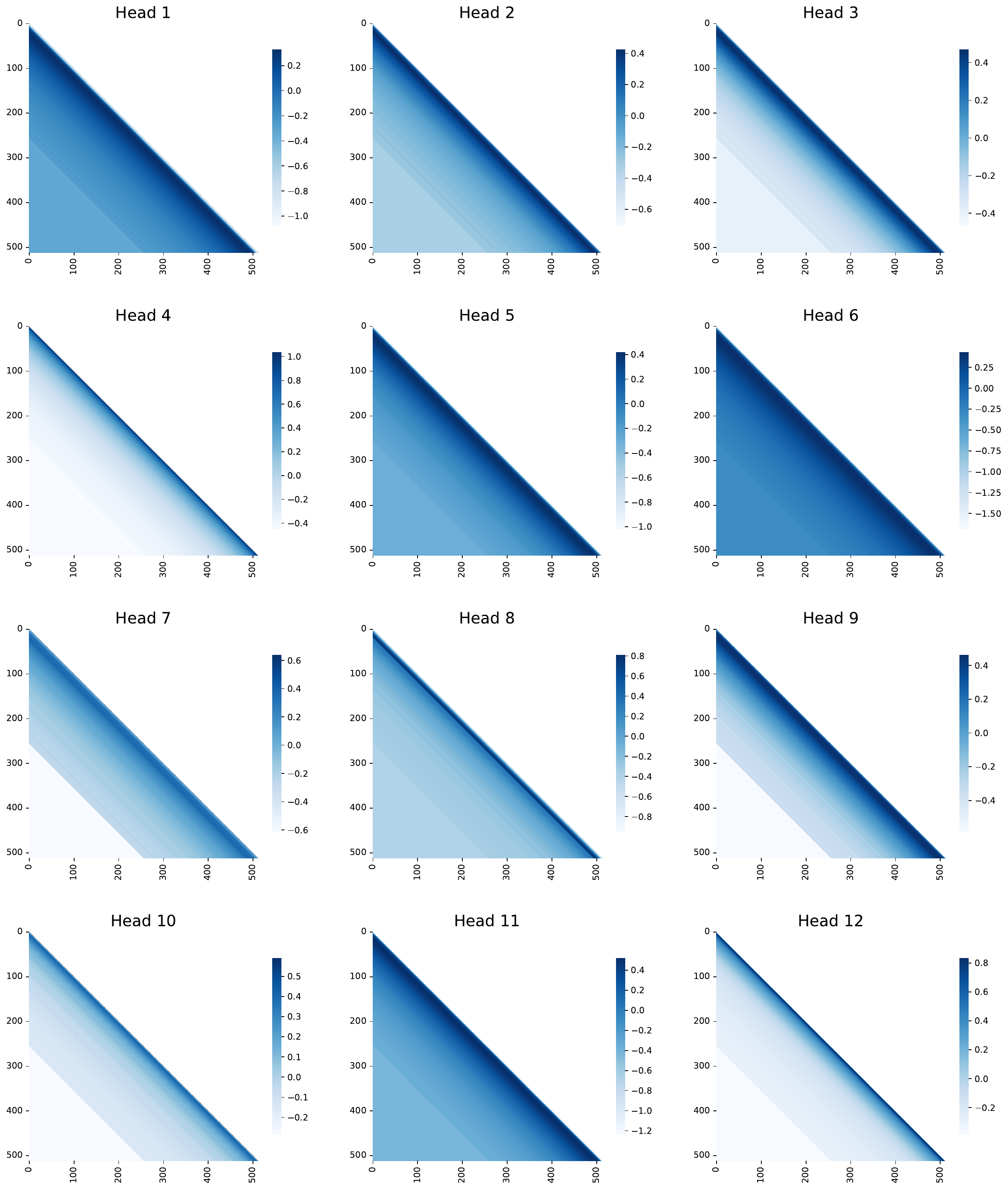}
    \end{adjustbox}
    \caption{Visualization of \textbf{T5-bias} positional biases across all attention heads in the final layer of GPT-2 Small, shown for a randomly sampled input sequence. As like other additive RPEs, these biases are added to the attention scores. The uniform patterns across tokens highlight the context-independent nature of T5-bias.}
    \label{fig:all_heads_t5bias}
\end{figure*}

\begin{figure*}[!ht]
    \centering
    \begin{adjustbox}{max width=\linewidth}
        \includegraphics{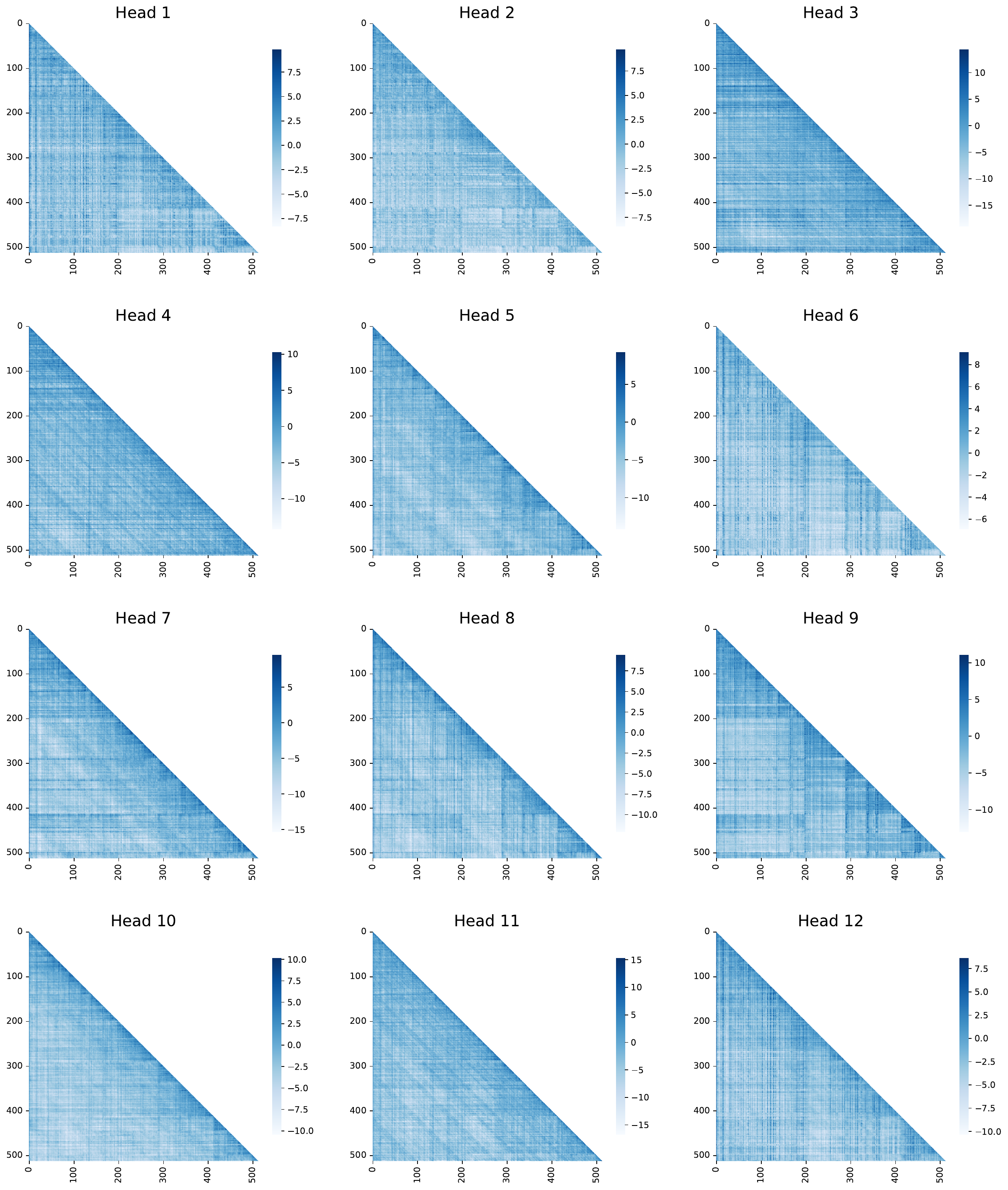}
    \end{adjustbox}
    \caption{Visualization of \textbf{RoPE} attention scores across all attention heads in the final layer of GPT-2 Small, shown for a randomly sampled input sequence. Since, RoPE is a non-additive RPE, these scores are the final attention scores, obtained by modifying the query and key vectors.}
    \label{fig:all_heads_rope}
\end{figure*}

Figures~\ref{fig:all_heads_cable} to \ref{fig:all_heads_t5bias} illustrate the additive biases incorporated into the attention scores ($B$ in Equation~\ref{eq:additiveRPE}) for the additive RPEs Cable, DAPEv2+Cable, Kerple, DAPEv2+Kerple, ALiBi, Fire, and T5-bias. The visualizations are derived from the final layer of GPT-2 Small model using a randomly selected input example. Since RoPE does not introduce an explicit additive bias but instead modifies the query and key representations, the corresponding attention scores after this transformation are shown separately in Figure~\ref{fig:all_heads_rope}.

\end{document}